\documentclass{article}

    \usepackage[preprint]{neurips_2020}

\usepackage[utf8]{inputenc} 
\usepackage[T1]{fontenc}    
\usepackage{hyperref}       
\usepackage{url}            
\usepackage{booktabs}       
\usepackage{amsfonts}       
\usepackage{nicefrac}       
\usepackage{microtype}      

\usepackage{hyperref}
\usepackage{amsmath}
\usepackage{mathtools}
\usepackage{multirow}
\usepackage{array}
\usepackage{siunitx}
\usepackage{booktabs}
\usepackage{algorithm}
\usepackage{graphicx}
\usepackage{booktabs} 

\usepackage{times}
\usepackage{epsfig}
\usepackage{graphicx}
\usepackage{amsmath}
\usepackage{amssymb}
\usepackage{algorithmic}
\usepackage{caption}
\usepackage{kpfonts}
\usepackage{comment}
\usepackage{xcolor}
\usepackage{ulem}
\usepackage{subfig}

\title{Dynamic Spatiotemporal Graph Neural Network with Tensor Network}

\author{%
  Chengcheng Jia, Bo Wu, Xiao-Ping Zhang\thanks{Use footnote for providing further information
    about author (webpage, alternative address)---\emph{not} for acknowledging
    funding agencies.} \\
  Department of Electrical, Computer $\&$ Biomedical Engineering \\
  Ryerson University, Toronto, ON M5B 2K3 \\
  \texttt{cc.jia@ryerson.ca, bowu1004@gmail.com, xzhang@ee.ryerson.ca} \\
}

\begin{document}

\maketitle

\begin{abstract}



Dynamic spatial graph construction is a challenge in graph neural network (GNN) for time series data problems.
Although some adaptive graphs are conceivable, only a 2D graph is embedded in the network to reflect the current spatial relation, regardless of all the previous situations.
In this work, we generate a spatial tensor graph (STG) to collect all the dynamic spatial relations, as well as a temporal tensor graph (TTG) to find the latent pattern along time at each node. 
These two tensor graphs share the same nodes and edges, which leading us to explore their entangled correlations by Projected Entangled Pair States (PEPS) to optimize the two graphs.
We experimentally compare the accuracy and time costing with the state-of-the-art GNN based methods on the public traffic datasets.

\end{abstract}

\section{Introduction}
\label{submission}


Recently, Graph Neural Network (GNN) based methods are widely used for computer vision tasks, such as image/video classification, Neural Language processing (NLP), time series prediction \citep{zhou2018graph} etc. This method constructs a graph from a dataset by exploiting the local correlation among the data, then embeds the graph into the neural network to improve the performance.
There are mainly four categories of GNNs, which are recurrent GNNs, convolutional GNNs,  graph Auto-Encoders, and spatial-temporal GNNs \citep{wu2019comprehensive}. This paper focuses on the last GNN type.

There are two challenges in the time series prediction task with spatial-temporal GNNs methods.
The first challenge is how to construct a dynamic spatial graph. The current GNN method
\citep{yu2017spatio} uses a static spatial graph and 1D convolution on the temporal direction for traffic prediction, and this graph doesn't reflect the dynamic changing nodes and edges.
However, during the time changing, some nodes and edges may appear or disappear, which results in the graph changing. Therefore, it is reasonable to construct the dynamic graph to reflect the changing correlations.
Accordingly, Graph WaveNet \citep{wu2019graph} constructs an adaptive adjacency matrix to learn the spatial dependencies of all the nodes, then the model sums up all the time steps of convolutions with the adaptive matrices, but it doesn't reflect the changing graphs in a progressive manner.
To overcome this, we construct a dynamic spatial graph in a tensor structure, stacking all the temporal changing graphs together for convolution. This tensor graph is supposed to show the appearance and disappearance of nodes and edges along time, therefore could extract more fine features by high-dimensional convolution to improve the performance.

\begin{figure}[t]
\begin{center}
\centerline{\includegraphics[width=0.5\columnwidth]{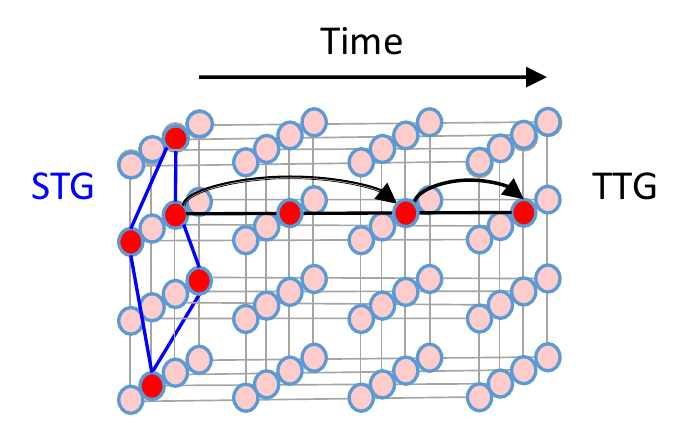}}
\caption{STG and TTG modeling. Each node connects an STG and a TTG in a time series dataset.}
\label{graph}
\end{center}
\end{figure}

Second, we aim to explore the latent temporal correlation of each node. The current popular GNN methods exploit only spatial structure of dataset to construct the graph, but the latent temporal correlation is neglected usually. Taking traffic prediction for example, the traffic is different along time at each station, like the rush hour in the morning and afternoon, therefore different time steps containing the similar traffic can be considered have stronger connection than those with obvious different traffics. In this paper, we construct a temporal graph along time for each node, to help to improve the performance by the spatial graph only.
In a word, to solve the two challenges, we construct not only a Spatial Tensor Graph (STG) but also a Temporal Tensor Graph (TTG), to catch both the spatial correlation among nodes and the temporal correlation for each node. The two graphs are shown in Figure~\ref{graph}.

Considering both the STG and TTG sharing all the nodes, i.e., an STG and TTG are ''entangled`` in the proposed model, we use a tensor network Projected Entangled Pair States (PEPS) \citep{verstraete2004renormalization} to optimize both the graphs simultaneously.
PEPS is a kind of Tensor Network (TN), which does tensor contraction with small ranks, i.e., reduced number of parameters, but with higher computational complexity \citep{cichocki2014era}.

In this paper, we propose a Dynamic Spatiotemporal GNN method with Tensor Network (DSTGNN) for traffic forecasting. 
The proposed method has three main contributions. 
\begin{itemize}
\item Two graphs, spatial and temporal graphs, are constructed to capture both the spatial correlation and the temporal correlation from one dataset, then are embedded into a GNN based model, to improve the performance of time series tasks.
\item Either STG or TTG contains dynamic information, and is constructed in a 3D tensor manner. This tensor structure of graph could reflect the appearance or disappearance of nodes and edges progressively, in order to extract finer features.
\item Considering STG and TTG constructed from one dataset, we use PEPS to optimize them simultaneously. Besides, PEPS can reduce the number of parameters of the two graphs by tensor contraction.
\end{itemize}

\section{Methodology}
In this section, we first describe the problem. Next, we give the details of construction of STG and TTG, the Spatial/Temporal Graph Convolution Layer (STGCL), and the network architecture. Last, we introduce PEPS in our model.
\subsection{Problem Definition}
We aim to predict traffic using GNN based proposed method. A dynamic graph is given as $G=(V,E,t)$, $V$ and $E$ are nodes (stations) and edges sets respectively at time $t$. 
Dataset is represented as $\mathcal{X}\in\mathbb{R}^{b\times T\times N\times D}$, while $b$ is the number of samples, $T$ is time steps, $N$ is the number of nodes, and $D$ is the dimension.
Given time duration $[1,l]$, our goal is to find a function $f$ to predict the next $T'$ data of $\mathcal{X}[:,1:l,::]$, which are $\mathcal{X}[:,l+1:l+T',::]$.

\subsection{Spatial Tensor Graph and Temporal Tensor Graph}

\subsubsection{Spatial Tensor Graph (STG)}

In our model, we construct STG using number of samples in all stations.
STG is represented as $\mathcal{A}\in\mathbb{R}^{N\times N\times T}$, where each element $\mathcal{A}[i,j,t]$ ($i,j\in[1,N], t\in[1,T]$) is 
\begin{equation*}
    \mathcal{A}[i,j,t]=
    \begin{cases} 
      \textup{exp}(-\frac{d_{ijt}^2}{\sigma^2}) &, i\neq j\hspace{1mm}\textup{and}\hspace{1mm} \textup{exp}(-\frac{d_{ijt}^2}{\sigma^2}) \geq \epsilon \\
      0 &, \textup{otherwise},
        \end{cases}
\end{equation*}
where $\mathcal{A}[i,j,t]$ is the weight of edge from $d_{ijt}$ (the subtract of samples between station $i$ and $j$ at time $t$), $\sigma^2$ and $\epsilon$ are threshold set to be $0.1$ and $0.5$, respectively.
In order to calculate the dynamic graph STG, we initialize graph at time $0$, which is $\mathcal{A}[::0]$. We perform Singular Value Decomposition (SVD) on the graph at each step to calculate $E_1, E_2$, then do ReLU and SoftMax for normalization as \citep{wu2019graph} does. The different of our construction of dynamic graph is we update the graph at time $t$ by $t-1$, we don't update the STG in each iteration, for less time costing. The details are in Algorithm~\ref{alg:stg}.

\begin{figure}[t]
\begin{center}
\centerline{\includegraphics[width=0.5\columnwidth]{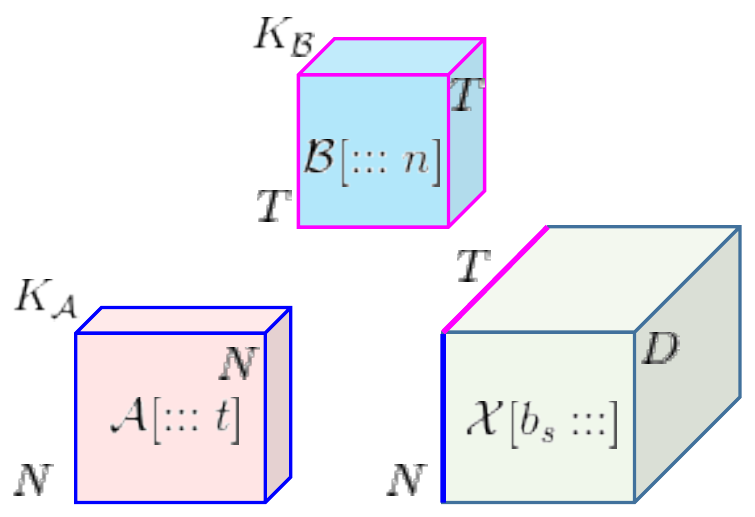}}
\caption{STGCL illustration. $\mathcal{X}[b_s:::]$ is a sample, $\mathcal{A}[:::t]$ is a spatial graph on time $t$ performed on mode-1 of dataset, and $\mathcal{B}[:::n]$ is a temporal graph for node $n$ performed on mode-2.}
\label{stgcl}
\end{center}
\end{figure}

\begin{algorithm}[tb]
   \caption{STG ($\mathcal{A}$) Construction}
   \label{alg:stg}
\begin{algorithmic}
   \STATE {\bfseries Input:} Graph $\mathcal{A}[::0]\in\mathbb{R}^{N\times N\times T}$
   \FOR{$t=1$ {\bfseries to} $T$}
   \STATE $E_1, E_2\leftarrow \textup{SVD}(\mathcal{A}^{(t-1)})$
      \STATE $\Delta\mathcal{A}^{(t-1)} = \textup{SoftMax}(\textup{ReLU}(E_1E_2^T))$
   \STATE $\mathcal{A}^{(t)}\leftarrow \mathcal{A}^{(t-1)} + \Delta\mathcal{A}^{(t-1)}$
   \ENDFOR
\end{algorithmic}
\end{algorithm}

\subsubsection{Temporal Tensor Graph (TTG)}
In our model, TTG is reprsented as $\mathcal{B}\in\mathbb{R}^{T\times T\times N}$, where $\mathcal{B}[t_1,t_2,n]$ is the absolute subtract of samples of time $t_1$ and $t_2$ at node $n$ ($t_1,t_2\in[1,T],n\in[1,N]$).
The construction of TTG is similar with that of STG, we omit the details for saving space.

\subsubsection{Chebyshev Polynomials Approximation}
One way of GNN is spectral graph convolution, which reduces computational time from quadratic to linear \citep{defferrard2016convolutional}.
Given a kernel $\Theta$, convolution operator $*_\mathcal{G}$, and signal $x$, the spectral graph convoluton is defined as $\Theta*_\mathcal{G}x=\Theta(L)x=\Theta(U\Lambda U^T)x$, where $U\in\mathbb{R}^{N\times N}$ is the matrix of eigenvectors of graph Laplacian matrix $L=I_N-D^{-\frac{1}{2}}A D^{-\frac{1}{2}}=U\Lambda U^T\in\mathbb{R}^{N\times N}$, where $D$ is the diagonal matrix of node degrees, $D_{ii}=\sum_{j}(A_{ij})$, $A\in\mathbb{R}^{N\times N}$ is the adjacency matrix of graph, $i,j\in[1,N]$.
Given a Chebyshev polynomial $T_k(\hat{L})\in\mathbb{R}^{N\times N}$ of order $k$, the graph convolution is rewritten as
\begin{equation}
\Theta *_\mathcal{G} x=\Theta(L)x \simeq\sum_{k=0}^{K-1}\theta_kT_k(\hat{L})x,
\end{equation}
where Laplacian matrix $\hat{L}=2L/\lambda_{max}-I_N$.
Here we use Chebyshev approximation to transform previous 3D graph to 4D $\mathcal{A}\in\mathbb{R}^{N\times N\times K_\mathcal{A} \times T}$ and $\mathcal{B}\in\mathbb{R}^{T\times T\times K_\mathcal{B} \times N}$, where $K_\mathcal{A}$ is kernel size of $\mathcal{A}$, the same as $K_\mathcal{B}$.

\subsection{Spatial/Temporal Graph Convolution Layer (STGCL)}
Given dataset $\mathcal{X}\in\mathbb{R}^{b,T,N,D}$ as a 4D tensor, the STG $\mathcal{A}$ is used on the mode-1 of $\mathcal{X}$, and the TTG $\mathcal{B}$ for mode-2 of $\mathcal{X}$. The illustration of this layer is shown in Figure~\ref{stgcl}. There are two layers in our model: Spatial Graph Convolution Layer (SGCL) and Temporal Graph Convolution Layer (TGCL). STGCL is designed as 
\begin{equation}
    \mathcal{X}_{out}=\underbrace{\mathcal{B}\mathcal{X}\mathcal{W}_\mathcal{B}}_{\textup{TGCL}}+\underbrace{\mathcal{A}\mathcal{X}\mathcal{W}_\mathcal{A}}_{\textup{SGCL}} + \underbrace{\mathcal{B}\mathcal{X}\mathcal{W}_\mathcal{B}}_{\textup{TGCL}},
\label{xout}
\end{equation}
where $\mathcal{W}_\mathcal{A}\in\mathbb{R}^{C_i K_\mathcal{A} \times C_o \times T}$, $\mathcal{W}_\mathcal{B}\in\mathbb{R}^{C'_i K_\mathcal{B} \times C'_o \times N}$, $C_i,C'_i$ are input dimension, $C_o,C'_o$ are output dimension. The TGCL and SGCL are introduced in Algorithm~\ref{alg:stgcl}.

\begin{figure}[t]
\begin{center}
\centerline{\includegraphics[width=0.7\columnwidth]{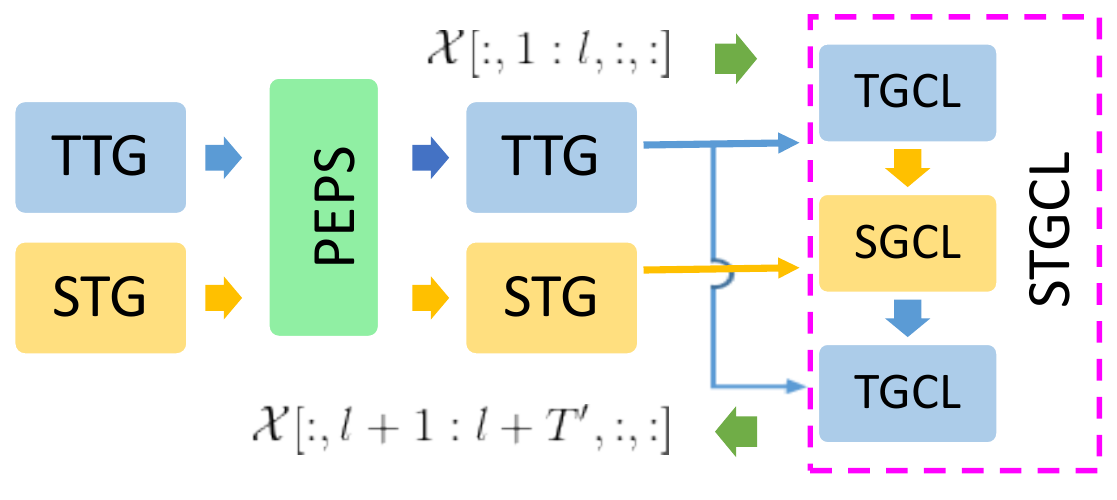}}
\caption{Framework illustration. The right part is the main framework of STGCL, with input STG and TTG. The left part is pre-processing of STG and TTG by PEPS, to reduce parameters with lower rank of the two graphs.}
\label{framework}
\end{center}
\end{figure}

\begin{algorithm}[tb]
  \caption{Spatial/Temporal Graph Convolution Layer}
  \label{alg:stgcl}
\begin{algorithmic}
  \STATE {\bfseries Input:} Data $\mathcal{X}\in\mathbb{R}^{b\times T\times N\times D}$, \\STG $\mathcal{A}\in\mathbb{R}^{N\times N\times K_\mathcal{A}\times T}$, TTG $\mathcal{B}\in\mathbb{R}^{T\times T\times K_\mathcal{B}\times N}$
  \STATE {\bfseries Output:} Data $\mathcal{X}\in\mathbb{R}^{b\times T\times N\times D}$

  \FOR{$n=1$ {\bfseries to} $N$}
      \STATE $\mathcal{X}\leftarrow \mathcal{B}[:::n]\mathcal{X}[::n:]\mathcal{W}_\mathcal{B}$\hspace*{3em}%
      \rlap{\smash{$\left.\begin{array}{@{}c@{}}\\{}\\{}\end{array}\color{magenta}\right\}%
      \color{magenta}\begin{tabular}{l}TGCL\end{tabular}$}}
  \ENDFOR

  \FOR{$t=1$ {\bfseries to} $T$}
  \STATE $\mathcal{X}\leftarrow \mathcal{A}[:::t]\mathcal{X}[:t::]\mathcal{W}_\mathcal{A}$\hspace*{3.4em}%
    \rlap{\smash{$\left.\begin{array}{@{}c@{}}\\{}\\{}\end{array}\color{blue}\right\}%
      \color{blue}\begin{tabular}{l}SGCL\end{tabular}$}}
  \ENDFOR
\end{algorithmic}
\end{algorithm}

\subsection{Network Architecture}
The framework is shown in Figure~\ref{framework}. The input of the model is dataset $\mathcal{X}[:,1:l,:,:]$, graphs STG and TTG as kernels, and the output is predictive data $\mathcal{X}[:,l+1:l+T',:,:]$ in the future $T'$ steps. The loss function is defined as
\begin{equation}
\begin{split}
        L(f;\Phi)=
        \|f(\mathcal{X}[:,1:l,::],\Phi)-
        \mathcal{X}[:,l+1:l+T',::]\|^2,
\end{split}
\end{equation}
where $\Phi=\{\mathcal{A},\mathcal{B},\mathcal{W}_\mathcal{A},\mathcal{W}_\mathcal{B}\}$. 

\begin{figure}[t]
\begin{center}
\centerline{\includegraphics[width=0.4\columnwidth]{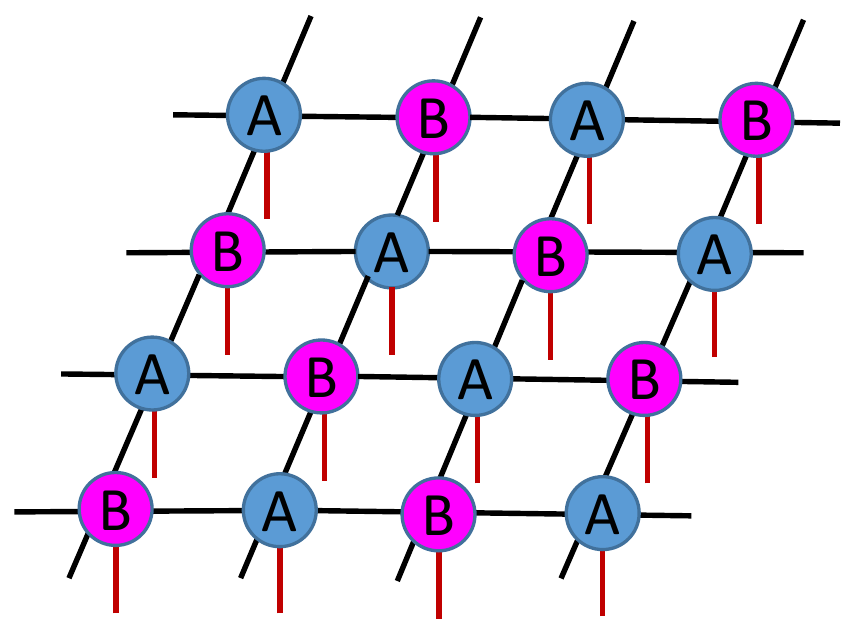}}
\caption{PEPS illustration. $A$ and $B$ are replaced by STG and TTG in our model.}
\label{peps}
\end{center}
\end{figure}

\subsection{Projected Entangled Pair States (PEPS)}

The PEPS structure is shown in Figure~\ref{peps}. We use PEPS for pre-precessing of STG and TTG in our framework, shown in Figure~\ref{framework}.

\subsection{Discussion}

\subsubsection{Tucker in STGCL}
In our model, STGCL can be designed in a Tucker decomposition manner, and Eq(\ref{xout}) is rewritten as
\begin{equation}
\begin{split}
    \mathcal{X}_{out}&=\Big[[\mathcal{X}\times_1\mathcal{A}]\mathcal{W}_\mathcal{A}\times_2\mathcal{B}\Big]\mathcal{W}_\mathcal{B}\\
    &=\mathcal{X}\times_1[\mathcal{A}\mathcal{W}_\mathcal{A}]\times_2[\mathcal{B}\mathcal{W}_\mathcal{B}]\\
    &=[\mathcal{X}\times_1\mathcal{A}\times_2\mathcal{B}]\mathcal{W}_\mathcal{A}\mathcal{W}_\mathcal{B},
\end{split}
    \label{tucker}
\end{equation}
while the difference from STGCL is to integrate SGCL and TGCL into one layer. In this case, High Order Singular Value Decomposition (HOSVD) could be used to obtain $\mathcal{W}_\mathcal{A}$ and $\mathcal{W}_\mathcal{B}$ to reduce computational complexity, instead of two directional convolution.

\subsubsection{PEPS effect}
PEPS here has two advantages, first, it digs out the correlation between STG $\mathcal{A}$ and TTG $\mathcal{B}$, since they are established from the same dataset.
Second, it reduces parameters of the two graphs by low rank learning.

\vspace{-4mm}
\section{Experiments}

\begin{table*}[t]
\caption{Performance comparison on METR-LA data set.}
\label{result}
\vskip 0.15in
\begin{center}
\begin{small}
\begin{sc}
\setlength\tabcolsep{1.5pt}
\begin{tabular}{lllll|lll|lllr}
\toprule
\multirow{2}{*}{Data} & \multirow{2}{*}{Models} & \multicolumn{3}{c}{15 min} & \multicolumn{3}{c}{30 min} &\multicolumn{3}{c}{60 min}\\\cline{3-11}
&&MAE &RMSE&MAPE&MAE &RMSE&MAP&MAE &RMSE&MAP\\
\midrule
\parbox[t]{0.1mm}{\multirow{10}{*}{\rotatebox[origin=c]{90}{METR-LA}}}
&ARIMA \citep{li2017diffusion}   & 3.99&8.21&9.60\%&5.15&10.45&12.70\%&6.90&13.23&17.40\% \\
&FC-LSTM \citep{li2017diffusion} &3.44&6.30&9.60\%&3.77&7.23&10.90\%&4.37&8.69&13.20\%\\
&WaveNet \citep{oord2016wavenet}    &2.99&5.89&8.04\%&3.59&7.28&10.25\%&4.45&8.93&13.62\% \\
&DCRNN \citep{li2017diffusion}    &2.77&5.38&7.30\%&3.15&6.45&8.80\%&3.60&7.60&10.50\%        \\
&GGRU \citep{zhang2018gaan}     &2.71&5.24&6.99\%&3.12&6.36&8.56\%&3.64&7.65&10.62\%\\
&STGCN \citep{yu2017spatio}      & 5.53&7.91&8.89\%&5.84&8.39&9.40\%&6.35&9.09& 10.23\%\\
&Graph WaveNet \citep{wu2019graph}      & \textbf{2.69}&5.15 &6.90\% &3.07 &6.22 &8.37\% &{3.53} &7.37 &10.01\%         \\
& DSTGNN [STG]  &5.46&7.87&8.79\%&5.50 &7.95&8.85\%&5.57&8.17&8.97\% \\
& DSTGNN [STG+TTG]  &5.03&7.41&8.10\% &5.14&7.52&8.27\% &5.15&8.12&8.29\% \\
& DSTGNN [STG+TTG+PEPS]  &2.75&\textbf{4.13}&\textbf{4.43}\% &\textbf{2.82}&\textbf{4.31}&\textbf{4.54}\% &\textbf{3.41}&\textbf{5.05}&\textbf{5.49}\% \\
\bottomrule
\end{tabular}
\end{sc}
\end{small}
\end{center}
\vskip -0.1in
\end{table*}

\subsection{Dataset}
METR-LA contains four months of statistics on traffic speed over $39,000$ sensor stations, covering the areas of California state highway system \citep{chen2001freeway}. This dataset is aggregated into $5$-minute interval from $30$-second data samples. We select District $7$ of California containing $256$ stations, labeled METR-LA. The time range is January and April of $2012$. We split the training and testing sets based on \citep{wu2019graph}.
\subsection{Baselines}
\begin{itemize}
    \item ARIMA. Auto-Regressive Intergrated Moving Average model with Kalman filter \citep{li2017diffusion}.
    \item FC-LSTM. Recurrent neural network with fully connected LSTM hidden units \citep{li2017diffusion}.
    \item WaveNet. A convolution network architecture for sequence data \citep{oord2016wavenet}.
    \item DCRNN. Diffusion convolution recurrent neural network \citep{li2017diffusion}, which combines graph convolution networks with recurrent neural networks in an Encoder-Decoder manner.
    \item GGRU. Graph gated recurrent unit network \citep{zhang2018gaan}, which is a recurrent-based approaches. GGRU uses attention mechanisms in graph convolution.
    \item STGCN. Spatial-temporal graph convolution network \citep{yu2017spatio}, which combines  graph convolution with 1D convolution.
    \item Graph WaveNet. Graph neural network \citep{wu2019graph} using an adaptive adjacency matrix of graph, and WaveNet performed on the temporal information.
\end{itemize}
\subsection{Experimental Setups}
Our experiments are conducted under a computer environment with one Intel(R) Core(TM) i7 CPU $920$ $@$ $2.67$ GHz and one NVIDIA Titan Xp GPU card.
The evaluation metrics are Mean Absolute Error (MAE), Root Mean Squared Error (RMSE), and Mean Absolute Percentage Error (MAPE).
\subsection{Results}

Table~\ref{result} shows the results of DSTGNN under STG, STG$+$TTG, and STG$+$TTG integrated with PEPS situations.
\vspace{-3mm}


\section{Conclusion}
In this paper, we propose a novel GNN framework with two graphs, STG and TTG, to capture both spatial and temporal local changing information. Both two graphs are dynamic constructed progressively. Experiemnts show that this two graph structure achieves better performance than one graph. Moreover, we exploit PEPS to optimize the two graphs, to reduce parameters by lower rank.

\clearpage
\bibliographystyle{unsrtnat}
\bibliography{neurips_2020}

\end{document}